\documentclass[runningheads]{llncs}

\usepackage{eccv}

\usepackage{eccvabbrv}

\usepackage{graphicx}
\usepackage{booktabs}

\usepackage[accsupp]{axessibility}  

\usepackage{wrapfig}
\usepackage{graphicx}
\usepackage{amsmath}
\usepackage{amssymb}
\usepackage{booktabs}
\usepackage{mathtools}
\usepackage{iac_pkg}
\usepackage{lipsum}
\usepackage{amssymb}
\usepackage{pifont}
\usepackage{multirow}
\usepackage{tabularx, booktabs}
\usepackage[htt]{hyphenat}
\usepackage{color}
\usepackage{xspace}
\usepackage{cite}
\usepackage{overpic}
\usepackage{arydshln}
\usepackage{subcaption}
\usepackage{xcolor}
\usepackage{bm}

\definecolor{citecolor}{RGB}{34,139,34}
\definecolor{lightred}{RGB}{255,100,100}
\definecolor{cell_bisque}{rgb}{1.0, 0.89, 0.77}
\definecolor{cell_blond}{rgb}{0.98, 0.94, 0.75}
\definecolor{cell_blue}{RGB}{155, 187, 228}
\definecolor{princetonorange}{rgb}{1.0, 0.56, 0.0}
\definecolor{pinkpearl}{rgb}{0.91, 0.67, 0.81}
\definecolor{mossgreen}{rgb}{0.68, 0.87, 0.68}

\newcommand{\Paragraph}[1]{\vspace{1mm}\noindent\textbf{#1.}\hspace{0mm}}
\newcommand{\Section}[1]{\vspace{0mm} \section{#1} \vspace{0mm}}
\newcommand{\SubSection}[1]{\vspace{0mm} \subsection{#1} \vspace{0mm}}

\usepackage[pagebackref,breaklinks,colorlinks,citecolor=eccvblue]{hyperref}

\usepackage{orcidlink}

\begin{document}

\title{GLARE: Low Light Image Enhancement via Generative Latent Feature based Codebook Retrieval} 

\titlerunning{GLARE}



\author{Han Zhou\inst{1,*}\orcidlink{0000-0001-7650-0755} \and
Wei Dong\inst{1,*}\orcidlink{0000-0001-6109-5099} \and
Xiaohong Liu\inst{2,\dagger}\orcidlink{0000-0001-6377-4730} \and
Shuaicheng Liu\inst{3}\orcidlink{0000-0002-8815-5335} \and
\\Xiongkuo Min\inst{2}\orcidlink{0000-0001-5693-0416} \and
Guangtao Zhai\inst{2}\orcidlink{0000-0001-8165-9322} \and
Jun Chen\inst{1,\dagger}\orcidlink{0000-0002-8084-9332}}


\authorrunning{H.~Zhou et al.}


\institute{$^1$ McMaster University, $^2$ Shanghai Jiao Tong University, \\ $^3$ University of Electronic Science and Technology of China \\
\email{\{zhouh115, dongw22, chenjun\}@mcmaster.ca} \quad \email{liushuaicheng@uestc.edu.cn} \\
\email{\{xiaohongliu, minxiongkuo, zhaiguangtao\}@sjtu.edu.cn} \\ 
$^*$Equal Contribution \quad \quad $^{\dagger}$Corresponding Authors}

\maketitle

\begin{abstract}
Most existing Low-light Image Enhancement (LLIE) methods either directly map Low-Light (LL) to Normal-Light (NL) images or use semantic or illumination maps as guides. However, the ill-posed nature of LLIE and the difficulty of semantic retrieval from impaired inputs limit these methods, especially in extremely low-light conditions. To address this issue, we present a new LLIE network via \textbf{G}enerative  \textbf{LA}tent feature based codebook \textbf{RE}trieval (\textbf{GLARE}), in which the codebook prior is derived from undegraded NL images using a Vector Quantization (VQ) strategy. More importantly, we develop a generative Invertible Latent Normalizing Flow (I-LNF) module to align the LL feature distribution to NL latent representations, guaranteeing the correct code retrieval in the codebook. In addition, a novel Adaptive Feature Transformation (AFT) module, featuring an adjustable function for users and comprising an Adaptive Mix-up Block (AMB) along with a dual-decoder architecture, is devised to further enhance fidelity while preserving the realistic details provided by codebook prior. Extensive experiments confirm the superior performance of GLARE on various benchmark datasets and real-world data. Its effectiveness as a preprocessing tool in low-light object detection tasks further validates GLARE for high-level vision applications. Code is released at \url{https://github.com/LowLevelAI/GLARE}.
\end{abstract}

\section{Introduction}
\label{sec:intro}

Low light images often suffer from various degradations, including loss of details, reduced contrast, amplified sensor noise, and color distortion, making many downstream tasks challenging, such as object detection, segmentation, or tracking~\cite{tracking2016, tracking2021tip, detection2022eccv, segmentation, segmentation2}. Consequently, LLIE has been extensively studied recently. 

\begin{figure}[t]

  \centering
  \begin{subfigure}[h]{1\linewidth}
  \centering
    \includegraphics[width=\linewidth]{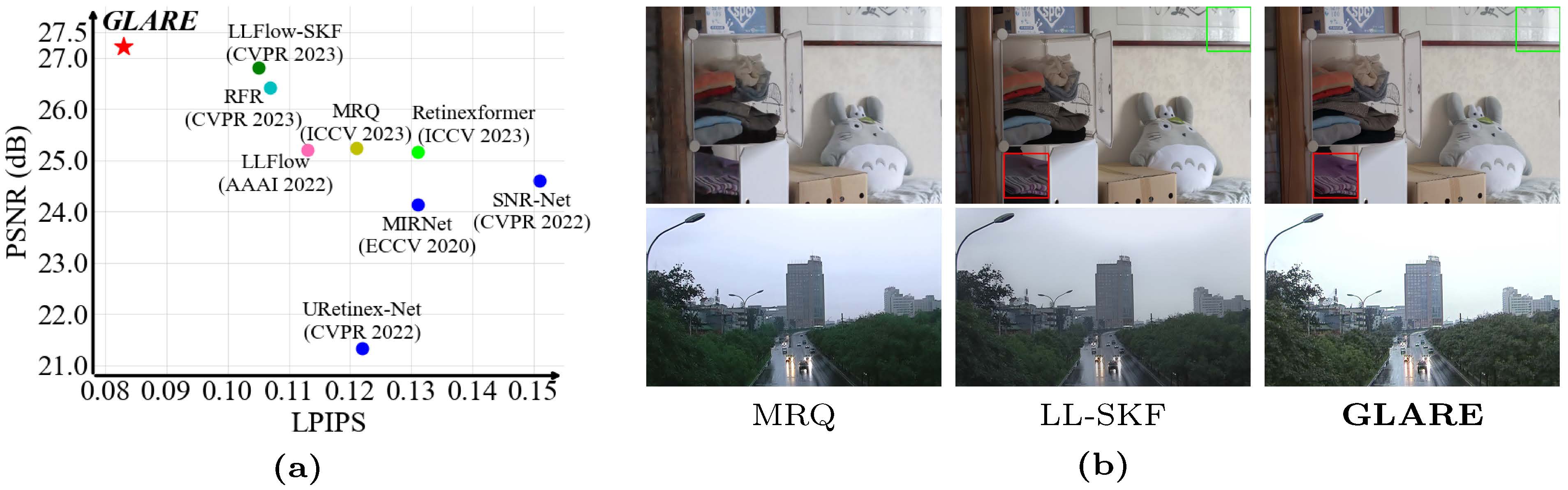}
  \end{subfigure}
  
\caption{(a) GLARE significantly outperforms SOTA methods on LOL~\cite{LOLv1}. (b) GLARE can generate appealing results on both LOL (upper) and real-world (lower) images.}
\label{fig_compare_first}

\end{figure} 

Traditional techniques that leverage handcrafted priors and constraints~\cite{Arici2009hismodi, Ibrahim2009dynamichis, fu2016variationalmodel, Jobsob1997multiretinex, zheng2022semanticzeroshot} have made significant contributions to this field. However, these methods still exhibit limitations in terms of adaptability across diverse illumination scenarios~\cite{illumin2019cvpr}. With the rapid advancements in deep learning, extensive approaches have been employed to learn complex mappings from LL to NL images~\cite{22, drbn2020cvpr,llnet2017,contrastenhencer2018}. Although their performance surpasses that of traditional methods, once deployed in real-world scenarios with varying light conditions and significant noise, these methods tend to produce visually unsatisfactory results.

Recent methods utilize semantic priors~\cite{semantic2023cvpr,zheng2022semanticzeroshot, issr2020acmmm}, extracted feature~\cite{smg2023cvpr, edgeguide2021, edgeiccv2021, edgeaaai2020}, and illumination maps~\cite{illumin2019cvpr} as the guidance to tackle the uncertainty and ambiguity of ill-posed LLIE problem.
However, they still face challenges in extracting stable features from heavily degraded inputs, which are often overwhelmed by noise and obfuscated by low visibility. Besides, only utilizing the extracted information from degraded images to build the LL-NL transformation usually generate unsatisfactory results when testing on real-world scenarios.

To generate realistic and appealing outcomes, one possible solution is to exploit the prior knowledge of natural normal-light images. Therefore, we propose to leverage a learned Vector-Quantized (VQ) codebook prior that captures the intrinsic features of high-quality and well-lit images, to guide the learning of LL-NL mapping. The discrete codebook is learned from noise-free images via VQGAN~\cite{Esser2021taming}. It is worth noting that by projecting degraded images onto this confined discrete prior space, the ambiguity inherent in LL-NL transformation is substantially mitigated, thereby ensuring the quality of enhanced images. Fig.~\ref{fig_compare_first} 
illustrates the superiority of our method over current state-of-the-art (SOTA) methods, both on the benchmark dataset and real-world images. 

\begin{figure}[ht]
  \centering
  \begin{subfigure}[h]{0.295\linewidth}
  \centering
    \includegraphics[width=\linewidth]{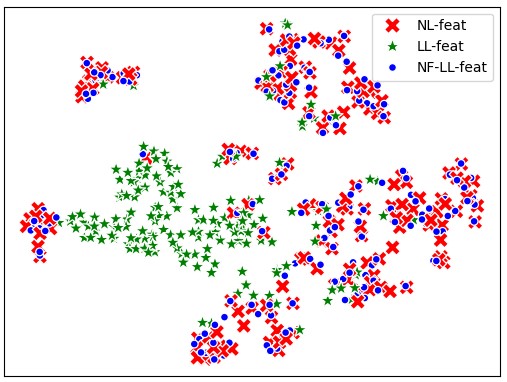}
    \caption{}
    \label{fig_second_dis}
  \end{subfigure}
  \hfill
  \begin{subfigure}[h]{0.695\linewidth}
  \centering
    \includegraphics[width=\linewidth]{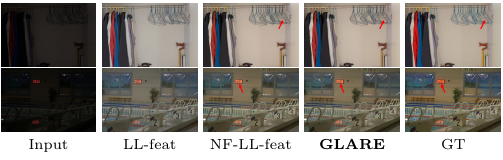}
	
  \caption{}
  \label{fig_second_vis}
  \end{subfigure}
  
\caption{(a) T-SNE~\cite{t-SNE} visualization of distributions of NL features, LL features, and LL-NF features. Compared to LL features, NF-LL features are better aligned with NL features. (b) Visual observations on each stage of GLARE on LOL~\cite{LOLv1} dataset. Column $2$-$4$ represent the enhanced results from LL-feat, the enhanced images from NF-LL-feat, and the final results of our \textbf{GLARE}. From column $2$-$4$, we observe a noticeable improvement on visibility, color preservation and detail recovery, which demonstrates the effectiveness of each stage of our GLARE. [Key: NL-feat: generated by NL encoder with NL inputs, LL-feat: LL features obtained form the fine-tuned NL encoder (Stage I), NF-LL-feat: LL features generated by our generative I-LNF module (Stage II)]}
\label{fig_second}

\end{figure}

It is important to emphasize that the superior performance of GLARE over other SOTA methods is not only attributed to the integration of the codebook prior but also to our unique designs that address two main challenges associated with leveraging the codebook prior for LL-NL mapping. First, as shown in Fig.~\ref{fig_second_vis} column $2$, solely exploiting VQGAN and NL prior may lead to unpleasant results and the reason behind this lies in the evident discrepancy between the degraded LL features and NL features in latent space, as depicted in Fig.~\ref{fig_second_dis}. Since the Nearest Neighbor (NN) is commonly utilized in looking up codebook~\cite{31,30,32,35}, this misalignment poses challenges in accurately retrieving VQ codes for LLIE task. Second, we notice that relying solely on matched codes for feature decoding~\cite{35,32,30} might compromise the fine details. Without integrating information from the original LL input, it could potentially introduce texture distortions.

Taking into account these issues, we further introduce two specific modules into GLARE. First, to bridge the gap between degradation features and NL representations, we introduce a generative strategy to produce LL features based on Invertible Latent conditional Normalizing Flow (I-LNF), which enables better alignment with potentially matched NL features. Specifically, given LL-NL pairs, our I-LNF transforms complicated NL features into a simple distribution with the condition of LL features via the precise log-likelihood training strategy. As shown in Fig.~\ref{fig_second_dis}, through this fully invertible network, our GLARE achieves a generative derivation of LL features which are closely aligned with NL representations and ensures accurate code assembly in codebook, thereby generating better enhancement results as depicted in Fig.~\ref{fig_second_vis} column $3$.  

Second, to improve the texture details, we propose an Adaptive Feature Transformation (AFT) module equipped with learnable coefficients to effectively control the ratio of encoder features introduced to the decoder. By flexibly merging the LL information into the decoding process, our model exhibits resilience against severe image degradation and one can freely adjust these coefficients according to their preference for real-world image enhancement. Besides, the AFT module adopts a dual-decoder strategy, which includes the fixed Normal-Light Decoder (NLD) and the trainable Multi-scale Fusion Decoder (MFD). The NLD specifically processes matched codes from the codebook, facilitating the generation of realistic and natural results. Meanwhile, the MFD handles the LL features produced by our I-LNF module, enhancing the final result with more refined details and texture, as demonstrated in Fig.~\ref{fig_second_vis} column $4$.

\Paragraph{Contributions} The main contributions of this work are as follows: \textbf{(i)} We are the first to adopt the external NL codebook as a guidance to enhance low-light images naturally. \textbf{(ii)} We introduce \textbf{GLARE}, a novel LLIE enhancer leveraging latent normalizing flow to learn the LL feature distribution that aligns with NL features. \textbf{(iii)} A novel adaptive feature transformation module with an adjustable function for users is proposed to consolidate the fidelity while ensuring the naturalness in outputs. \textbf{(iv)} Extensive experiments indicate that our method significantly outperforms existing SOTA methods on $5$ paired benchmarks and $4$ real-world datasets in LLIE and our model is highly competitive while employed as a pre-processing method for high-level object detection task.

\section{Related Work}
\label{sec:related}
\subsection {Deep Learning based LLIE methods} 
Similar to numerous approaches in other image restoration tasks~\cite{dehazedct2024cvprw, shadowremoval2024cvprw, GridDehazeNet, GridDehazeNet+, FMSNet,NTIRE_Dehazing_2024, vasluianu2024ntire_isr}, end-to-end LLIE methods~\cite{llnet2017, MIRNet, contrastenhencer2018, decom2018cvpr, enlighengan2021, FastLLVE, AttentionLut} have been proposed to directly map LL images to NL ones. Most of them mainly resort to the optimization of reconstruction error between the enhanced output and ground-truth to guide the network training. However, they often fail to preserve naturalness and restore intricate details effectively. These problems have given rise to the exploration of leveraging additional information or guidance to aid the enhancement process. For instance, some methods~\cite{19, illumin2019cvpr} achieve a simple training process for LLIE by estimating illumination maps. However, these approaches have a risk of amplifying noise and color deviations especially in real-world LL images~\cite{retinexformer}.

Concurrently, other methods~\cite{issr2020acmmm, zheng2022semanticzeroshot, semantic2023cvpr} argue that semantic understanding can mitigate regional degradation problems and attain pleasing visual appearance. Besides, several studies ~\cite{edgeguide2021, edgeaaai2020, smg2023cvpr} have indicated that utilizing edge extraction can direct the generation of realistic image details and mitigate the blurry effects to an extent. Nevertheless, these methods are highly contingent upon features extracted from degraded input, which could compromise the generalization capability and introduce artifacts. In contrast to existing methods, we propose an informative codebook that encapsulates a diverse spectrum of NL feature priors. This approach demonstrates resilience against various degradations, achieving more natural and appealing image enhancement.

\begin{figure*}[ht]
    \centering
    \includegraphics[width=\linewidth]{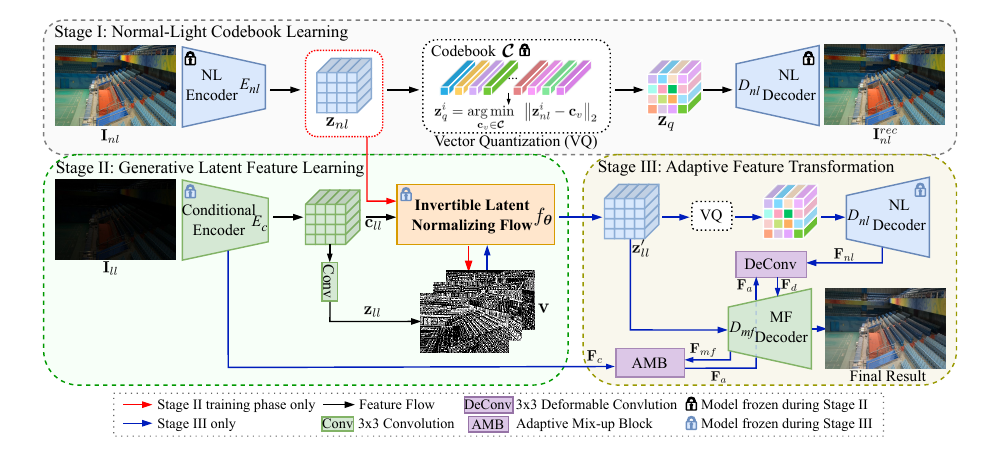}
    \caption{The overall architecture of our proposed \textbf{GLARE}. There are three training stages in our model. Stage I aims to learn a comprehensive normal-light codebook $\bm{\mathcal{C}}$ using VQGAN. In Stage II training, given the $\mbf{c}_{ll}$ and $\mbf{z}_{ll}$ generated by the conditional encoder $E_{c}$ and the convolution layer respectively, I-LNF module $f_{\bm{\theta}}$ learns to transform the normal-light feature $\mbf{z}_{nl}$ to a simplified Gaussian distribution $\mbf{v} = f_{\bm{\theta}}(\mbf{z}_{nl}; \mbf{c}_{ll})$ with the mean of $\mbf{z}_{ll}$. We optimize $ E_{c}$ and $f_{\bm{\theta}}$ by minimizing the negative log-likelihood described in Eq.~\ref{eq_nll}. In Stage III, the codebook $\bm{\mathcal{C}}$, the NL decoder $ D_{nl}$, the conditional encoder $ E_{c}$, and the I-LNF $f_{\bm{\theta}}$ are all fixed. Our GLARE can effectively transform a Gaussian density $ p_{\bm{v}}(\mbf{v})\sim \mathcal{N}(\mbf{z}_{ll}, \bm{\Sigma})$ to the NL feature distribution $p_{\mbf{z}_{nl}|\mbf{c}_{ll} }(\mbf{z}_{nl}|\mbf{c}_{ll}, \bm{\theta}) $. To further improve the enhancement performance, we propose an adaptive feature transformation strategy. By leveraging $D_{mf}$ and AMB to flexibly incorporate LL information for decoding ($\mbf{F}_d = DeConv(\mathbf{F}_{nl}, AMB(\mbf{F}_c, \mbf{F}_{mf}))$), our GLARE is capable to generate results with more refined texture and details. [Key: $\bm{\Sigma}$: The unit variance]}

    \label{fig_third}
\end{figure*}

\subsection{Vector-Quantized Codebook Learning} 
Vector Quantized Variational AutoEncoders (VQ-VAE) is firstly proposed in ~\cite{28} to learn discrete representations. This approach effectively tackles the posterior collapse issue that is commonly encountered in VAE models. Then, VQVAE2~\cite{33} explores the hierarchical VQ code for large-scale image generation. VQGAN~\cite{Esser2021taming} further enhances the perceptual quality by capturing a codebook of context-rich visual parts via an adversarial method. The discrete codebook has been successfully employed in image super-resolution~\cite{30}, text super-resolution~\cite{34}, and face restoration~\cite{31,32,36}. However, there remains potential for further improvement. One of the key research directions is how to precisely match the related correct code. Different from recent methods~\cite{36,37} that utilize a Transformer to predict code indices in the codebook, we argue that prediction-based strategies are inherently unable to address the significant differences between LL and NL features, resulting in suboptimal performance. To this end, we propose a generative approach that produces LL features aligned with NL counterparts to successfully bridge the gap between LL and NL representations.

\section{GLARE}
\label{sec:method}
Besides introducing external NL codebbok to guide the Low-Light to Normal-Light (LL-NL) mapping, the novelty of our work also lies in the distinctive Invertible Latent Normalizing Flow (I-LNF) and Adaptive Feature Transformation (AFT) modules, which are designed to maximize the potential of NL codebook prior and generate realistic results with high fidelity. The overview of our method is illustrated in Fig.~\ref{fig_third}, where the training of our method can be divided into three stages. In stage I, we pre-train a VQGAN on thousands of clear NL images to construct a comprehensive VQ codebook (Sec.~\ref{sec_codebook_dict}). In stage II, the I-LNF module is trained utilizing LL-NL pairs to achieve the distribution transformation between LL and NL features (Sec.~\ref{sec_normalization_flow}). In the final stage, the AFT module, which contains the fixed NL Decoder (NLD), Adaptive Mix-up Block (AMB), and Multi-scale Fusion Decoder (MFD), is employed to enhance the fine-grained details while maintaining naturalness beneficial from the codebook (Sec.~\ref{sec_adptive_fusion}).

\SubSection{Stage I: Normal-Light Codebook Learning } \label{sec_codebook_dict}
To learn a universal and comprehensive codebook prior, we leverage a VQGAN with the structure similar to~\cite{Esser2021taming}. Specifically, a NL image $ \mbf{I}_{nl} \in \mathbb{R}^{3 \PLH W \PLH H}$ is first encoded and reshaped into the latent representation $ \mbf{z}_{nl} \in \mathbb{R}^{d \PLH N}$, where $W$, $H$, $d$, and $N= W/f \PLH H/f$ represent the image width, image height, the dimension of latent features, and the total number of latent features; $f$ is the downsampling factor of the NL Encoder $E_{nl}$. Each latent vector $\mbf{z}^{i}_{nl}$ can be quantized to the corresponding code $\mbf{z}^{i}_{q}$ using Nearest-Neighbor (NN) matching as:

\begin{equation}
 \mbf{z}^{i}_{q} =\mathop {\arg\min}\limits_{\mbf{c}_{v} \in \bm{\mathcal{C}}} \ \ \left \|  \mbf{z}^{i}_{nl} - {\mbf{c}_{v}} \right \|_2,
\label{eq_z_q}
\end{equation}
where $\bm{\mathcal{C}} \in \mathbb{R}^{d \PLH N_{c}}$ denotes the learnable codebook containing $N_{c}$ discrete codes, with each element represented by $\mbf{c}_{v}$. Then the quantized code $ \mbf{z}_{q}$ is sent to NLD (denoted as $D_{nl}$) to generate reconstructed image $ \mbf{I}^{rec}_{nl}$. 

To better illustrate the strengths and limitations of the codebook prior, we fine-tune the pre-trained VQGAN encoder on LL-NL pairs. Specifically, we achieve the enhanced results shown in Fig.~\ref{fig_second_vis} column $2$ and we utilize t-SNE~\cite{t-SNE} to visualize LL features generated by fine-tuned NL encoder in Fig.~\ref{fig_second_dis}, which demonstrate the effectiveness of external NL prior in IILE. Besides, these visual results inspire us to design additional networks to align LL features with NL representations to further improve enhancement performance.

\SubSection{Stage II: Generative Latent Feature Learning} \label{sec_normalization_flow}
To fully exploit the potential of external codebook prior, we design additional mechanisms from the perspective of reducing the disparity between LL and NL feature distributions. Specifically, we develop an invertible  latent normalizing flow to achieve the transformation between LL and NL feature distributions, thereby facilitating more accurate codes retrieval from codebook. 

As shown in Fig.~\ref{fig_third}, two key components are optimized in stage II: the \textbf{Conditional Encoder} and the \textbf{Invertible LNF}. 1) The conditional encoder $ E_{c}$, structurally identical to the NL encoder $E_{nl}$, inputs a LL image $\mbf{I}_{ll}$ and outputs the conditional feature $ \mbf{c}_{ll}$. 2) The I-LNF module in this work is realized through an invertible network, represented as $f_{\bm{\theta}}$. This module utilizes $\mbf{c}_{ll}$ as the condition to transform the complex NL feature distribution $\mbf{z}_{nl}$ to a latent feature, namely $\mbf{v} = f_{\bm{\theta}}(\mbf{z}_{nl}; \mbf{c}_{ll})$. Stage II focuses on obtaining a simplified distribution $p_{\mbf{v}}(\mbf{v})$ in the space of $\mbf{v}$, such as a Gaussian distribution. Consequently, the conditional distribution $p_{\mbf{z}_{nl}|\mbf{c}_{ll} }(\mbf{z}_{nl}|\mbf{c}_{ll}, \bm{\theta})$ can be implicitly expressed as~\cite{srflow2020}: 
\begin{equation}
p_{\mbf{z}_{nl}|\mbf{c}_{ll} }
(\mbf{z}_{nl}|\mbf{c}_{ll}, \bm{\theta})= p_{\mbf{v}} (f_{\bm{\theta}}(\mbf{z}_{nl}; \mbf{c}_{ll}))|det \frac{\partial f_{\bm{\theta}}}{\partial \mbf{z}_{nl}}(\mbf{z}_{nl}; \mbf{c}_{ll})|.    
\label{eq_conditional_dis}
\end{equation}

Different from conventional normalizing flow applications~\cite{srflow2020, llflow2022, semantic2023cvpr, nf2023iccv}, we uniquely employ normalizing flow at the feature level rather than the image space, and our I-LNF module is designed without integrating any squeeze layers. Moreover, instead of using the standard Gaussian distribution as the prior of $\mbf{v}$, we propose to use the LL feature $\mbf{z}_{ll} $, generated by convolution layers based on $ \mbf{c}_{ll} $, as the mean value of $p_{\mbf{v}} (\mbf{v})$. The conditional distribution in Eq.~(\ref{eq_conditional_dis}) allows us to minimize the negative log-likelihood (NLL) in Eq.~(\ref{eq_nll}) to train the conditional encoder and I-LNF module. Besides, through the fully invertible network $f_{\bm\theta}$, we derive clear features $ \mbf{z}^{\prime}_{ll} $ for LL inputs by sampling $\mbf{v}\sim p_{\bm{v}}(\mbf{v})$ according to $\mbf{z}^{\prime}_{ll} = f_{\mbf{\bm\theta}}^{-1}(\mbf{v};\mbf{c}_{ll})$. 

\begin{equation}
\mathcal{L}(\bm\theta; \mbf{c}_{ll}, \mbf{z}_{nl} ) = -\mathrm{log} p_{\mbf{z}_{nl}|\mbf{c}_{ll} }(\mbf{z}_{nl}|\mbf{c}_{ll}, \bm\theta).
\label{eq_nll}
\end{equation}

After training, we evaluate our model on LOL dataset~\cite{LOLv1} to validate the effectiveness of the I-LNF. As shown in Fig.~\ref{fig_second_dis}, the LL feature distribution generated by I-LNF is closely aligned with that of the NL, facilitating accurate code assembly in the codebook. Moreover, our network achieves satisfactory enhancement results (Fig.~\ref{fig_second_vis}, column $3$), indicating good LLIE performance after stage II. However, these results still exhibit considerable potential for improvement, especially in fidelity. For example, the color (row 1 in Fig.~\ref{fig_second_vis}) or structural details (row 2 in Fig.~\ref{fig_second_vis}) significantly diverge from the ground truth. This observation motivates and drives us to incorporate the input information into the decoding process to elevate the fidelity.

\SubSection{Stage III: Adaptive Feature Transformation} \label{sec_adptive_fusion}
To further enhance the texture details and fidelity, we propose an adaptive feature transformation module that flexibly incorporates the feature $\mbf{F}_{c} = \{\mbf{F}^{i}_{c}\}$ from the conditional encoder into the decoding process, where $i$ denotes the resolution level. Specifically, in order to maintain the realistic output of NLD and avoid the influence of degraded LL features, we adopt a dual-decoder architecture and develop MFD inspired by~\cite{ridcp2023, 31}. Dual-decoder design enables us to leverage the deformable convolution ($dconv$) to warp NLD feature ($\mbf{F}^{i}_{nl}$) as Eq.~\ref{eq_dconv} and input the warped feature ($\mbf{F}^{i}_{d}$) into MFD to generate the final enhancement.
\begin{equation}
\mbf{F}^{i}_{d} = dconv(\mbf{F}^{i}_{nl}, \mbf{F}^{i}_{t}),
\label{eq_dconv}
\end{equation}
where $i$ and $\mbf{F}^{i}_{t}$ denote the resolution level and the target feature respectively. In this work, we design a novel feature fusion network that adaptively incorporates LL information into the warping operation and provides a potential adjustment choice for users when testing on real-world occasions.

\Paragraph{Adaptive Mix-up Block} The MFD that aligns structurally with NLD aims to decode the generated LL feature $\mbf{z}^{\prime}_{ll}$ and obtain intermediate representations as $\mbf{F}_{mf}=\{\mbf{F}^{i}_{mf}\}$, where $i$ indicates the resolution level. At each resolution level, the conditional encoder information $\mbf{F}^{i}_{c}$ is added to the corresponding $\mbf{F}^{i}_{mf}$ in order to bring more LL information. Different from typical feature fusion operations (\textit{i.e.},~skip connection~\cite{vqiccv23}), our approach uses an adaptive mix-up strategy:
\begin{equation}
\mbf{F}^{i}_{a} = \beta \times \sigma(\theta_{i}) \times \mbf{F}^{i}_{c} + (1 - \beta \times \sigma(\theta_{i})) \times \mbf{F}^{i}_{mf},
\label{eq_adaptive_fusion}
\end{equation}
where $\theta_{i}$ represents a learnable coefficient, $\sigma$ denotes the sigmoid operator, $\beta$ is used for the adjustment for real-world testing and is set to $1$ when training. Unlike skip connection, these learnable parameters can be adjusted effectively during training, which contributes to enhanced performance shown in Sec.~\ref{sec_ex_abla}.

\Paragraph{Flexible Adjustment} Even though $\beta$ in Eq.~(\ref{eq_adaptive_fusion}) is set to $1$ in the training phase, one can flexibly adjust $\beta$ according to their preference when testing with real-world images. This design stems from the phenomena that many current methods usually work struggling with real-world data, which often have different illuminations with images used in the training phase.

\Section{Experiments}
\label{sec_ex}

\setlength\tabcolsep{1.5pt}
\begin{table}[t]
\caption{ Quantitative comparisons on LOL~\cite{LOLv1}, LOL-v2-real~\cite{LOLv2}, LOL-v$2$-synthetic~\cite{LOLv2}, SDSD-indoor~\cite{SDSD}, and SDSD-outdoor~\cite{SDSD} datasets. Our GLARE achieves superior performance compared to current SOTA methods. These results are obtained either from original papers or by running their released codes. [Key: \textbf{\red{Best}}, \blue{Second Best}, $\uparrow$ ($\downarrow$): Larger (smaller) values leads to better performance]. }
\centering

    \centering
        \resizebox{\linewidth}{!}{
        \begin{tabular}{c|ccc|ccc|cc|cc|cc}
        \hline
        \multirow{2}{*}{\begin{tabular}{c}
            \textbf{ Methods}
        \end{tabular}}&\multicolumn{3}{c|}{LOL-v$1$} &\multicolumn{3}{c|}{LOL-v$2$-real}&\multicolumn{2}{c|}{LOL-v$2$-syn}&\multicolumn{2}{c|}{SDSD-indoor}&\multicolumn{2}{c}{SDSD-outdoor}\\\cline{2-13}
        &PSNR$\uparrow$ &SSIM$\uparrow$ &LPIPS$\downarrow$ &PSNR$\uparrow$ &SSIM$\uparrow$ &LPIPS$\downarrow$ &PSNR$\uparrow$ &SSIM$\uparrow$ &PSNR$\uparrow$ &SSIM$\uparrow$ &PSNR$\uparrow$ &SSIM$\uparrow$\\ 
        \hline
        SID~\cite{SID}&$14.35$&$0.436$&---&$13.24$&$0.442$&---&$15.04$&$0.610$ &$23.29$&$0.703$&$24.90$&$0.693$\\
        IPT~\cite{IPT}&$16.27$&$0.504$&---&$19.80$&$0.813$&---&$18.30$&$0.811$ &$26.11$&$0.831$&$27.55$&$0.850$\\
        UFormer~\cite{Uformer}&$16.36$&$0.771$&---&$18.82$&$0.771$&---&$19.66$&$0.871$ &$23.17$&$0.859$&$23.85$&$0.748$\\
        Sparse~\cite{LOLv2}&$17.20$&$0.640$&---&$20.06$&$0.816$&---&$22.05$&$0.905$ &$23.25$&$0.863$&$25.28$&$0.804$\\
        RUAS~\cite{RUAS}&$18.23$&$0.720$&---&$18.37$&$0.723$&---&$16.55$&$0.652$ &$23.17$&$0.696$&$23.84$&$0.743$\\
        SCI~\cite{SCI}&$14.78$&$0.646$&---&$20.28$&$0.752$&---&$24.14$&$0.928$ &---&---&---&---\\
        KinD~\cite{KinD}&$20.87$&$0.802$&$0.207$&$17.54$&$0.669$&$0.375$&$13.29$&$0.578$ &$21.95$&$0.672$&$21.97$&$0.654$\\
        MIRNet~\cite{MIRNet}&$24.14$&$0.830$&$0.131$&$20.02$&$0.820$&$0.138$&$21.94$&$0.876$ &$24.38$&$0.864$&$27.13$&$0.837$\\
        DRBN~\cite{drbn2020cvpr}&$19.86$&$0.834$&$0.155$&$20.13$&$0.830$&$0.147$&$23.22$&$0.927$ &$24.08$&$0.868$&$25.77$&$0.841$\\
        SNR~\cite{SNR}&$24.61$&$0.842$&$0.151$&$21.48$&$0.849$&$0.157$&$24.14$&$0.928$ &$29.44$&$0.894$&$28.66$&$0.866$\\
        URetinex-Net~\cite{URetinex-Net}&$21.33$&$0.835$&$0.122$&$21.16$&$0.840$&$0.144$&$24.14$&$0.928$ &---&---&---&---\\
        Restormer~\cite{Restormer}&$22.43$&$0.823$&---&$19.94$&$0.827$&---&$21.41$&$0.830$ &$25.67$&$0.827$&$24.79$&$0.802$\\
        Retformer~\cite{retinexformer}&$25.16$&$0.845$&$0.131$&$22.80$&$0.840$&$0.171$&$25.67$&$0.930$ &$\blue{29.77}$&$\red{\mathbf{0.896}}$&$\blue{29.84}$&$\blue{0.877}$ \\
        MRQ~\cite{vqiccv23}&$25.24$&$0.855$&$0.121$&$22.37$&$0.846$&$0.142$&$25.94$&$0.935$ &---&---&---&--- \\
        LLFlow~\cite{llflow2022}&$25.19$&$0.870$&$0.113$&$26.53$&\blue{$0.892$}&$0.135$&$26.23$&$0.943$ &---&---&---&--- \\
        LL-SKF~\cite{semantic2023cvpr}&$\blue{26.80}$&$\blue{0.879}$&$\blue{0.105}$&$\blue{28.45}$&$\red{\mathbf{0.905}}$&$\blue{0.111}$&$\blue{29.11}$&$\blue{0.953}$ &---&---&---&--- \\
        \hline
        \textbf{GLARE (Ours)}&$\red{\mathbf{27.35}}$&$\red{\mathbf{0.883}}$&$\red{\mathbf{0.083}}$&$\red{\mathbf{28.98}}$&$\red{\mathbf{0.905}}$&$\red{\mathbf{0.097}}$&$\red{\mathbf{29.84}}$&$\red{\mathbf{0.958}}$ &$\red{\mathbf{30.10}}$&$\red{\mathbf{0.896}}$&$\red{\mathbf{30.85}}$&$\red{\mathbf{0.884}}$ \\
        \hline
        \end{tabular}
    } 

\label{table_quant_compar}

\end{table}
\setlength\tabcolsep{6pt}

\subsection{Datasets}
\label{sec_ex_datasets}
\Paragraph {Normal Light Datasets} To train the VQGAN in Stage I, we select images with normal lighting conditions from DIV2K~\cite{DIV} and Flickr2K~\cite{Flickr} datasets to develop the NL codebook prior.

\Paragraph {Low Light Datasets} 
We conduct a thorough evaluation of our method using various paired datasets, including LOL~\cite{LOLv1}, LOL-v2-real~\cite{LOLv2}, LOL-v2-synthetic \cite{LOLv2}, and a large-scale dataset SDSD~\cite{SDSD}. For LOL, LOL-v2-real, and LOL-v2-synthetic, we use 485, 689, and 900 pairs for training, and 15, 100, and 100 pairs for testing. The indoor subset of SDSD dataset includes 62 training and 6 testing video pairs, while the outdoor subset contains 116 training and 10 testing pairs. Besides, we also conduct cross-dataset evaluation on several unpaired real-world datasets: MEF~\cite{MEF}, LIME~\cite{LIME}, DICM~\cite{DICM}, and NPE~\cite{NPE}.

\subsection{Implementation Details} 
\label{sec_ex_imple} 
\Paragraph {Experiment Settings} We use the Adam optimizer ($\beta_{1}=0.9$, $\beta_{2}=0.99$) for all training stages. In Stage I, the training iteration is set to 640K with a batch size of $4$, a fixed learning rate $10^{-4}$, and image size of $256\times256$. In Stage II, we retain the batch size, change the image size to $320 \times 320$, and adopt a multi-stage learning rates. Then, our GLARE is trained for 60K iterations on LOL and LOL-v2, and 225K iterations on SDSD. In Stage III, the batch size is halved, the initial learning rate is set to $5\times 10^{-5}$, and the training iterations are adjusted to 20K, 40K, and 80K for LOL, LOL-v2, and SDSD datasets respectively.

\Paragraph {Metrics} For paired datasets, we utilize Peak Signal-to-Noise Ratio (PSNR) and Structural Similarity Index Measure (SSIM)~\cite{SSIM} to assess pixel-level accuracy, and use Learned Perceptual Image Patch Similarity (LPIPS)~\cite{LPIPS} for perceptual quality evaluation. As for real-world datasets, the Natural Image Quality Evaluator (NIQE)~\cite{NIQE} is adopted.

\subsection{Performance on LLIE}
\label{sec_ex_LLIE}
\Paragraph {Quantitative Results}
As reported in Tab.~\ref{table_quant_compar}, GLARE outperforms the current SOTA methods on five benchmarks. Our GLARE excels in PSNR, outperforming LL-SKF over 0.55 dB and 0.74 dB on LOL and LOL-v2-synthetic datasets. Furthermore, it surpasses Retformer with improvement of 0.33 dB and 1.01 dB on SDSD-indoor and SDSD-outdoor datasets. Additionally, our LPIPS scores surpass the second best performance by $\bm{20.9\%}$ and $\bm{12.6\%}$ in Tab.~\ref{table_quant_compar}, indicating that the enhanced results from our network are more consistent with human visual system. Tab.~\ref{table_unpaired} presents the cross-dataset evaluation on unpaired real-world datasets. We first train our GLARE on LOL training split. Then, the model that performs best on LOL test data is deployed on four unpaired datasets. As compared to the current SOTA methods, GLARE outperforms them on DICM and MEF and achieves the optimal performance on average. This demonstrates not only the superiority of our method in producing high-quality visual results but also its good generalization capability.

\begin{figure}[t]
    \centering 
	
    \includegraphics[width=\linewidth]{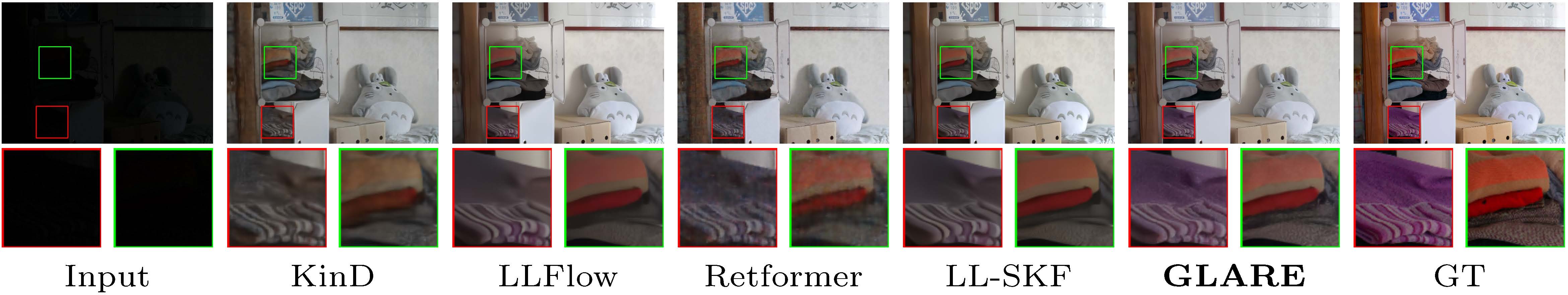}
	
    \caption{Visual comparisons on LOL~\cite{LOLv1} dataset. Our method can effectively enhance visibility and generate visually appealing results.}
    \label{fig:LOL-V1}
\end{figure}

\begin{figure}[t]
    \centering 
	
    \includegraphics[width=\linewidth]{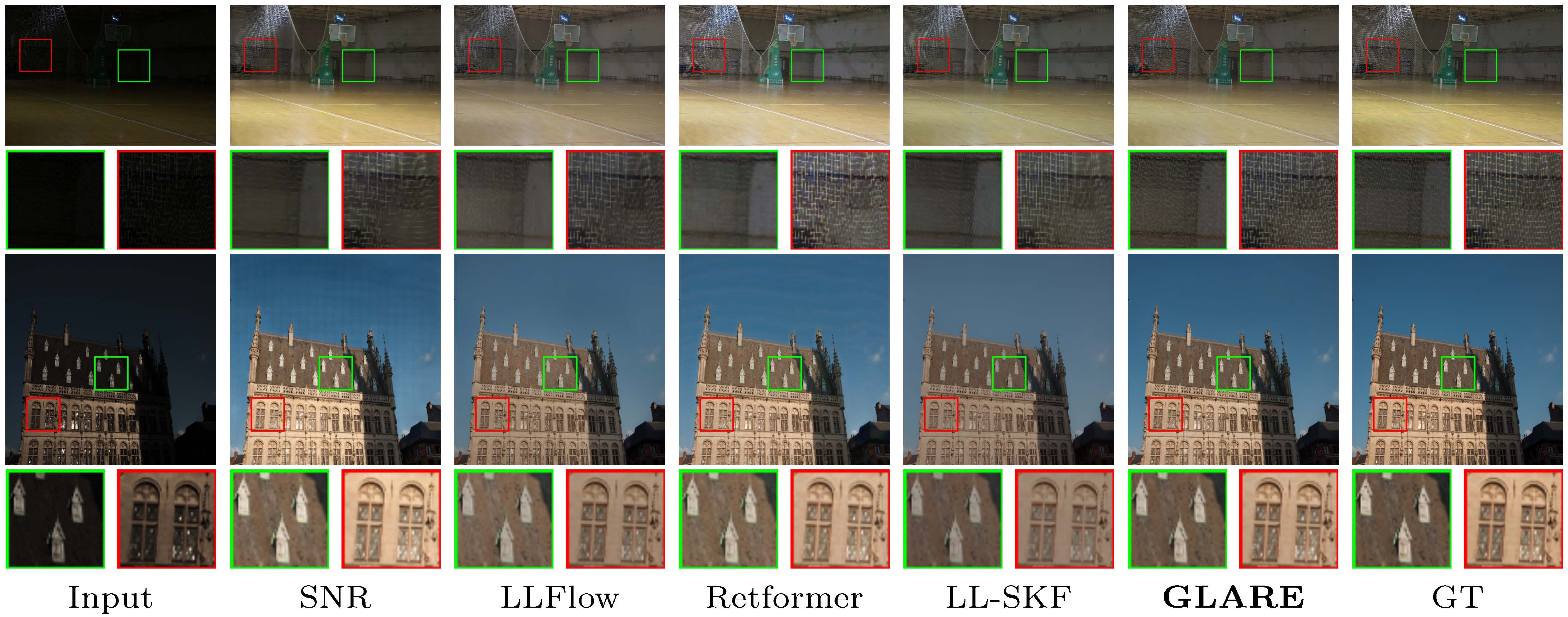}
	
    \caption{Visual comparisons on LOL-v2-real~\cite{LOLv2} (top) and LOL-v2-synthetic~\cite{LOLv2} (bottom) datasets. Previous methods suffer from either severe color distortion or detail deficiency, while our GLARE performs favorably without these issues.}
    \label{fig:LOL-V2}
\end{figure}

\setlength\tabcolsep{6pt}
\begin{table}[!t]
    \caption{ Quantitative comparisons on real-world datasets. These results are obtained either from the original papers or testing with their best LOL~\cite{LOLv1} weights. [Key: \textbf{\red{Best}}, \blue{Second Best}, $\downarrow$: Smaller value represents better quality].}
    \centering
    \scalebox{0.8}{
    \begin{tabular}{c|cccc|c}
    \hline
        
    \textbf{Methods}&MEF&LIME &DICM &NPE &Mean$\downarrow$  \\ 
    \hline
    SNR~\cite{SNR}&$4.14$&$5.51$&$4.62$&$4.36$&$4.60$\\
    URetinex-Net~\cite{URetinex-Net}&$\blue{3.79}$&$\blue{3.86}$&$4.15$&$4.69$&$4.11$\\
    LLFlow~\cite{llflow2022}&$3.92$&$5.29$&$3.78$&$4.16$&$3.98$\\
    LL-SKF~\cite{semantic2023cvpr}&$4.03$&$5.15$&$\blue{3.70}$&$\mathbf{\red{4.08}}$&$3.92$\\
    RFR~\cite{RFR}&$3.92$&$\mathbf{\red{3.81}}$&$3.75$&$\blue{4.13}$&$\blue{3.81}$\\
    \hline
    \bf{GLARE (Ours)}&$\mathbf{\red{3.66}}$&$4.52$&$\mathbf{\red{3.61}}$&$4.19$&$\mathbf{\red{3.75}}$\\
    \hline
    \end{tabular}
    }    

\label{table_unpaired}
\end{table}
\setlength\tabcolsep{6pt}

\begin{figure}[!t]
    \begin{minipage}[!t]{0.5\linewidth}
        \centering
         \begin{subfigure}[b]{1\linewidth}
              \centering
              \includegraphics[width=\linewidth, height=1.183\linewidth]{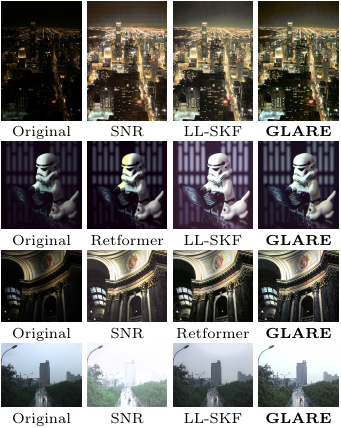}
        \end{subfigure}

        \caption{Visual results of cross-dataset evaluation on unpaired real-world datasets. These four images are from DICM~\cite{DICM}, LIME~\cite{LIME}, MEF~\cite{MEF}, and NPE~\cite{NPE} respectively. Our GLARE generates more pleasing results without noise or artifacts.}
        \label{fig:unpaired}
    \end{minipage}
    \hfill
\begin{minipage}[!t]{0.48\linewidth}

	\centering

	\begin{subfigure}[t]{1\linewidth}
		\centering
		\includegraphics[width=\linewidth]{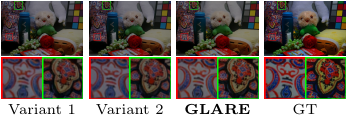}
	\end{subfigure}
        \caption{Visual ablation results of AFT on LOL~\cite{LOLv1}. Our GLARE with AFT module is capable to generate results with improved edge acuity and contour definition, along with a more abundant detail texture.}
        \label{fig:ablation-a}
\ \\        
       \begin{subfigure}[t]{1\linewidth}
		\centering
		\includegraphics[width=\linewidth]{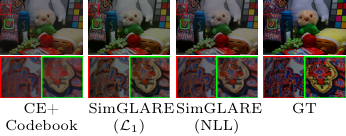}
	\end{subfigure}
        \caption{Visual comparisons for ablation study of I-LNF. Without our proposed I-LNF module, the results exhibit significant detail loss and blurriness, coupled with a notably darker tone in certain areas.}
        \label{fig:ablation-b}
    \end{minipage}
\end{figure}

\Paragraph {Qualitative Results}
The visual quality of our GLARE against others are shown in Fig.~\ref{fig:LOL-V1}, ~\ref{fig:LOL-V2}, and ~\ref{fig:unpaired}. Obviously, previous methods show inferior performance on noise suppression. Besides, they also tend to produce results with evident color distortion (See the enhanced results of KinD, LLFlow, Retformer, LL-SKF in Fig.~\ref{fig:LOL-V1}, and SNR, LLFlow in Fig.~\ref{fig:LOL-V2}). Additionally, from the qualitative comparison, it can be seen that LLFlow, LL-SKF, and Retformer may induce the detail deficiency on their enhanced results (See Fig.~\ref{fig:LOL-V1} and  Fig.~\ref{fig:LOL-V2}), and KinD in Fig.~\ref{fig:LOL-V1} and SNR in Fig.~\ref{fig:LOL-V2} perform poorly in vision due to the introduce of unnatural artifacts. In comparison, GLARE can effectively enhance poor visibility while reliably preserving color and texture details without artifacts. The visual comparisons on unpaired real world datasets in Fig.~\ref{fig:unpaired} also demonstrate the strengths of our method in terms of details recovery and color maintenance.

\SubSection{Performance on Low-Light Object Detection}
\label{sec_ex_OD}
\Paragraph{Implementation Details} To thoroughly evaluate our model, we also explore its potential as an effective preprocessing method in object detection task on ExDark dataset~\cite{Exdark}. This dataset collects 7,363 low-light images, categorized into 12 classes and annotated with bounding boxes. We first employ different LLIE models trained on LOL to enhance the ExDark dataset, then carry out object detection on the enhanced images. More concretely, 5,896 images are used for training and the rest for evaluation. The adopted object detector is  YOLO-v3~\cite{YOLO}  pre-trained on COCO dataset~\cite{COCO}.

\setlength\tabcolsep{1.5pt}
\begin{table}[t]
    \caption{Quantitative low-light detection results on ExDark~\cite{Exdark} using different LLIE method as the enhancement tool. [Key: \textbf{\red{Best}}, \blue{Second Best}, $\uparrow$: The larger represents the better performance, Baseline: These scores are obtained by training the YOLO-v3~\cite{YOLO} detector on original ExDark~\cite{Exdark} dataset].}
    \centering
    \resizebox{\linewidth}{!}{
    \begin{tabular}{c|cccccccccccc|c}
    \hline
        
    \textbf{Methods}&Bicycle &Boat &Bottle &Bus &Car &Cat &Chair &Cup &Dog &Motor &People &Table &Mean$\uparrow$  \\ 
    \hline
    Baseline~\cite{Exdark}&$80.4$&$76.5$&$\blue{77.6}$&$89.7$&$84.0$&$\blue{71.5}$&$69.5$&$76.4$&$78.7$&$76.4$&$\blue{81.9}$&$52.6$&$76.32$\\
    MBLLEN~\cite{MBLLEN}&$82.2$&$\mathbf{\red{77.5}}$&$76.3$&$90.3$&$84.1$&$70.9$&$69.4$&$75.9$&$77.7$&$74.7$&$\mathbf{\red{82.0}}$&$\mathbf{\red{58.2}}$&$76.59$\\
    KinD~\cite{KinD}&$79.7$&$\blue{77.4}$&$\mathbf{\red{78.8}}$&$\mathbf{\red{92.5}}$&$\mathbf{\red{84.9}}$&$70.8$&$67.5$&$78.3$&$78.7$&$\blue{77.1}$&$80.9$&$53.7$&$76.69$\\
    LLFlow~\cite{llflow2022}&$81.6$&$75.5$&$74.3$&$\mathbf{\red{92.5}}$&$84.5$&$69.7$&$69.0$&$75.8$&$\blue{79.0}$&$76.5$&$80.9$&$\blue{57.9}$&$76.44$\\
    IAT~\cite{IAT}&$\blue{82.5}$&$76.0$&$75.6$&$\blue{92.3}$&$83.0$&$\mathbf{\red{72.4}}$&$\blue{70.8}$&$\mathbf{\red{79.6}}$&$78.6$&$76.2$&$81.5$&$\blue{57.9}$&$\blue{77.19}$\\
    LL-SKF~\cite{semantic2023cvpr}&$80.2$&$75.0$&$76.6$&$91.3$&$\blue{84.7}$&$69.5$&$\mathbf{\red{71.1}}$&$76.5$&$77.5$&$76.4$&$81.1$&$57.1$&$76.43$\\
    \hline
    \bf{GLARE (Ours)}&$\mathbf{\red{83.4}}$&$75.8$&$\blue{77.6}$&$91.7$&$83.9$&$70.1$&$70.0$&$\blue{79.1}$&$\mathbf{\red{81.5}}$&$\mathbf{\red{77.2}}$&$\mathbf{\red{82.0}}$&$\blue{57.9}$&$\mathbf{\red{77.50}}$\\
    \hline
    \end{tabular}
    }    
\label{table_object}
\end{table}
\setlength\tabcolsep{6pt}

\begin{figure}[t]
    \centering 
	
    \includegraphics[width=\linewidth]{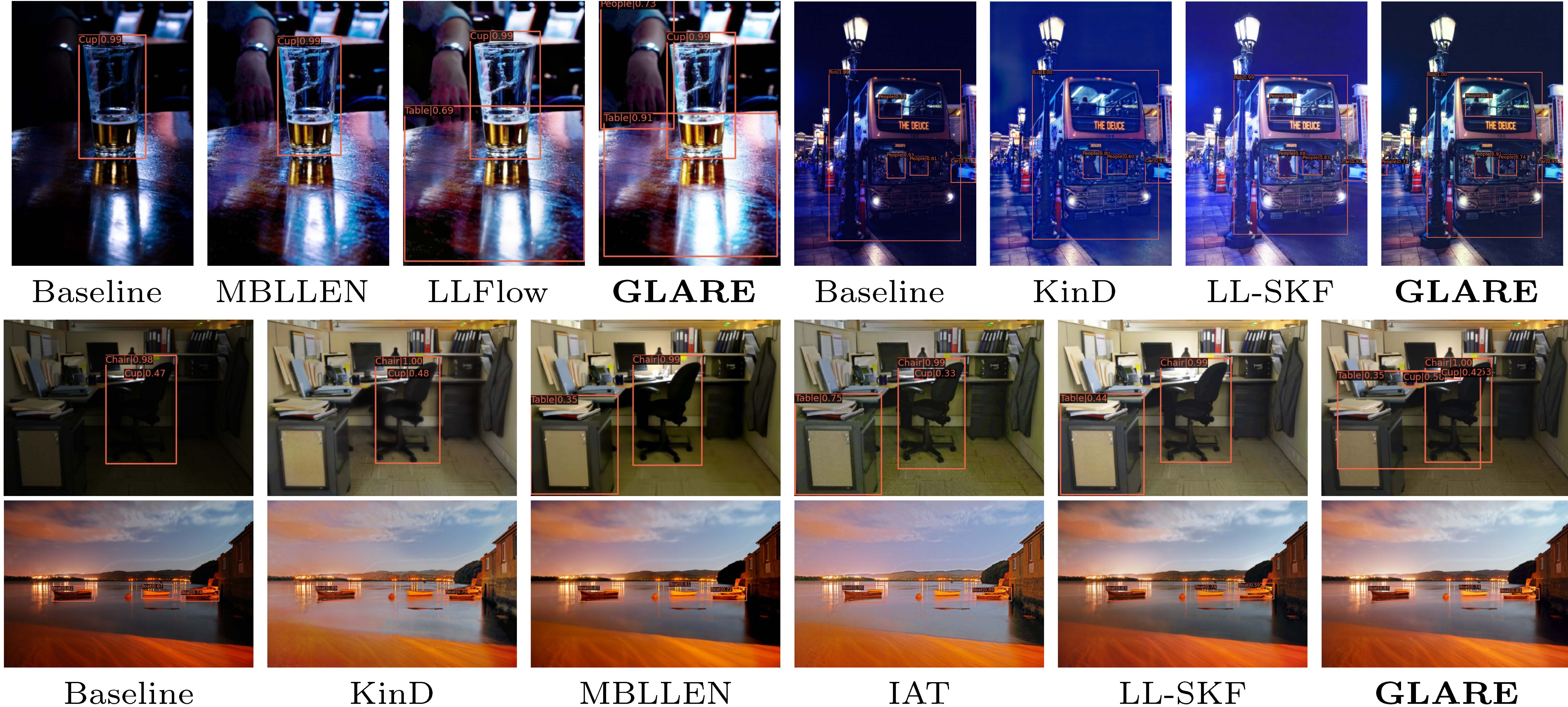}
	
    \caption{Visual comparisons and detection results of enhancement-based low-light object detection on the ExDark~\cite{Exdark} dataset. Previous enhancement methods, when employed as preprocessing modules for object detection, encountered issues with object loss. In contrast, utilizing images enhanced by our GLARE enables YOLO-v3 to robustly detect targets with high confidence and our enhanced images exhibit better visual quality. Please zoom in for better view. [Key: Baseline: Test results of YOLO-v3~\cite{YOLO} detector trained on original ExDark~\cite{Exdark} dataset].}
    \label{fig_visual_exdark}
\end{figure}

\Paragraph{Quantitative Results} We calculate the Average Precision (AP) and mean Average Precision (mAP) scores as our evaluation metrics. We compare our GLARE against current SOTAs in Tab.~\ref{table_object}. As compared to KinD, MBLLEN, LLFlow, and LL-SKF, our GLARE achieves at least 0.8 improvement in terms of mAP. More importantly, our GLARE also outperforms IAT, which has demonstrated exceptional performance in low-light object detection.

\Paragraph{Qualitative Results} The visual comparisons for low-light object detection is demonstrated in Fig.~\ref{fig_visual_exdark}. It can be seen that although each LLIE method enhances the visibility to some extent, GLARE achieves the best visual performance, thus benefiting the most to the downstream detection task. Not surprisingly, the enhanced results from GLARE enables the YOLO-v3 detector to recognize more objects with higher confidence.

\SubSection{Ablation Study}
\label{sec_ex_abla}

\setlength\tabcolsep{2pt}
\begin{table}[t]
\centering
\caption{By incorporating AFT, GLARE gains significant improvements on PSNR, SSIM, and LPIPS. Besides, our GLARE also performs better than two variants based on SimGLARE. [Key: SC: Skip Connection operation~\cite{concatation}, Dual-d: Dual decoder architecture, \red{\bf{Best}}, \blue{Second Best}] }.
\begin{subtable}[t]{1\linewidth}
    \centering
    
    \resizebox{0.7\linewidth}{!}{
    \begin{tabular}{c|ccc|ccc}
    \hline
        
    \textbf{Methods}&SC&AMB &Dual-d &PSNR$\uparrow$  &SSIM$\uparrow$  &LPIPS$\downarrow$  \\ 
    \hline
    (1) SimGLARE&&& &$26.51$&$0.855$&$0.109$\\
    (2) Variant 1&&\checkmark& &$26.60$&$0.867$&$0.093$\\
    (3) Variant 2&\checkmark&&\checkmark &$\blue{26.88}$&$\blue{0.877}$&$\blue{0.087}$\\
    (4) \bf{GLARE}&&\checkmark&\checkmark&$\red{\bf{27.35}}$&$\red{\bf{0.883}}$&$\red{\bf{0.083}}$\\
    \hline
    \textbf{Methods}&$\mathcal{L}_{1}$&$\mathcal{L}_{per}$&$\mathcal{L}_{ssim}$&PSNR$\uparrow$ &SSIM$\uparrow$ &LPIPS$\downarrow$ \\ 
    \hline
    (5) \bf{GLARE}&\checkmark&& &$27.02$&$0.870$&$0.099$\\
    (6) \bf{GLARE}&\checkmark&\checkmark& &$\blue{27.09}$&$\blue{0.871}$&$\blue{0.091}$\\
    (7) \bf{GLARE}&\checkmark&\checkmark&\checkmark&$\red{\bf{27.35}}$&$\red{\bf{0.883}}$&$\red{\bf{0.083}}$\\
    \hline
    \end{tabular}
    }    
\end{subtable}
\label{table_abla_aft}
\end{table}

\begin{table}[t]
\centering
\caption{NLL can help achieve better enhancement result compare to $\mathcal{L}_{1}$ loss. While replacing the I-LNF module with a transformer or removing I-LNF entirely, the significant performance decline highlights the importance of the proposed I-LNF module.}

\begin{subtable}[t]{1\linewidth}
    \centering
    \resizebox{0.75\linewidth}{!}{
    \begin{tabular}{c|ccc}
    \hline
        
    \textbf{Methods} &PSNR$\uparrow$  &SSIM$\uparrow$  &LPIPS$\downarrow$  \\ 
    \hline
    (1) SimGLARE ($\mathcal{L}_{1}$) &$\blue{25.69}$&$\blue{0.842}$&$\blue{0.132}$\\
    (2) Codebook + Transformer ($\mathcal{L}_{1}$) &$25.12$&$0.834$&$0.148$\\
    (3) Conditional Encoder + Codebook ($\mathcal{L}_{1}$)&$24.53$&$0.825$&$0.161$\\    
    (4) {SimGLARE} (NLL) &$\red{\bf{26.51}}$&$\red{\bf{0.855}}$&$\red{\bf{0.109}}$\\
    \hline
    \end{tabular}
    }    

\end{subtable}
\label{table_abla_lnf}
\end{table}

\begin{table}[t]
\centering
\caption{The quantitative results of ablation experiments related to NL codebook prior. We observe significant decrease on metrics when the codebook is removed.}

\begin{subtable}[t]{1\linewidth}
    \centering
    \resizebox{0.75\linewidth}{!}{
    \begin{tabular}{c|ccc}
    \hline
        
    \textbf{Methods} &PSNR$\uparrow$  &SSIM$\uparrow$  &LPIPS$\downarrow$  \\ 
    \hline
    (1) Encoder-Decoder ($\mathcal{L}_{1}$) &$22.79$&$0.804$&$0.195$\\
    (2) Conditional Encoder + Codebook ($\mathcal{L}_{1}$)&$\blue{24.53}$&$\blue{0.825}$&$\blue{0.161}$\\ 
    \hline
    (3) Variant 3 ($\mathcal{L}_{1}$) &$23.42$&$0.812$&$0.176$\\
    (4) {SimGLARE} ($\mathcal{L}_{1}$) &$\red{\bf{25.69}}$&$\red{\bf{0.838}}$&$\red{\bf{0.132}}$\\
    \hline
    \end{tabular}
    }    

\end{subtable}
\label{table_abla_codebook}
\end{table}
\setlength\tabcolsep{6pt}

To verify the effectiveness of each component of GLARE and  justify the optimization objective utilized for training, we conduct extensive ablation experiments on LOL dataset. Specifically, we discuss the importance of AFT module, I-LNF module, and NL codebook prior in this section.

\Paragraph {Effectiveness of Adaptive Feature Transformation} By removing the AFT module from our GLARE, we obtain a Simple LLIE model denoted as SimGLARE. Basically, SimGLARE only utilizes the information from NL codebook without feature transformation. The quantitative results of SimGLARE are shown in Tab.~\ref{table_abla_aft}. SimGLARE is quite competitive on LLIE in terms of PSNR, SSIM, and LPIPS (compared with SOTAs in Tab.~\ref{table_quant_compar}). However, with the proposed AFT module, our GLARE achieves further improvements on both quantitative metrics and visual results (as shown in Fig.~\ref{fig:ablation-a}). In addition,  various loss functions are examined in Tab.~\ref{table_abla_aft}, showing that our choice of losses  in Stage III is reasonable.

We also design two variants, named Variant 1 and Variant 2, to shed light on the importance of proposed dual-decoder architecture and AMB, respectively. Specifically, Variant 1 directly incorporates LL feature to NLD using AMB while Variant 2 adopts parallel decoder strategy but replaces AMB with skip connection operation~\cite{concatation}. By comparing (4) with (2) and (3) in Tab.~\ref{table_abla_aft}, we observe that PSNR and SSIM are negatively correlated with LPIPS, which verifies the effectiveness of our AMB and dual-decoder design. 

\Paragraph {Effectiveness of Invertible Latent Normalizing Flow} To show the importance of I-LNF and the adopted NLL loss, we implement several adaptations based on SimGLARE. (1) We train SimGLARE using $\mathcal{L}_{1}$ loss to validate the effectiveness of NLL loss  adopted in our work. (2) We replace the I-LNF module by leveraging a Transformer model structurally similar to~\cite{36} to directly predict the code index in the codebook. (3) We remove the I-LNF module in (1) and train the conditional encoder on LL-NL pairs. The quantitative results are reported in Tab.~\ref{table_abla_lnf}. The superiority of NLL loss can be verified by comparing (4) and (1). Moreover, a comparison between the images in the second and third columns of Fig.~\ref{fig:ablation-b} also reveals that the use of NLL loss, as opposed to $\mathcal{L}_{1}$ loss, results in sharper contours and edges. Besides, as compared to the Transformer-based code prediction, our proposed I-LNF module can help generate LL features that are better aligned with NL ones, thus ensuring more accurate code matching  and achieving superior performance. More importantly, with the I-LNF module removed from SimGLARE ($\mathcal{L}_{1}$), we notice a significant decrease  in PSNR (1.16 dB $\downarrow$) and SSIM (0.017 $\downarrow$), which demonstrates the effectiveness of our proposed I-LNF module. 

\Paragraph {Effectiveness of Codebook Prior} To investigate the importance of the NL codebook prior, based on SimGLARE ($\mathcal{L}_{1}$), we remove the codebook and the quantization process in VQGAN. The resulting model is denoted as Variant 3 and is trained using a strategy similar to that for SimGLARE ($\mathcal{L}_{1}$). Similarly, we remove the codebook in the model reported in row $3$ of Tab.~\ref{table_abla_lnf}  to learn the LL-NL mapping. Quantitative results are reported in Tab.~\ref{table_abla_codebook}. The absence of a codebook prior notably impacts performance, as evidenced by an average decrease of 2.0 dB in PSNR and a 0.024 drop in SSIM. This highlights the critical importance of the codebook prior in our method.

\section{Conclusion}
\label{sec:conclu}
A novel method named GLARE is proposed for LLIE.  In view of the uncertainty and ambiguity caused by ill-posed nature of LLIE, we leverage the normal light codebook, which is obtained from clear and well-exposed images using VQGAN,  to guide the LL-NL mapping. To better exploit the potential of codebook prior,  the invertible latent normalizing flow is adopted to generate LL features aligned with NL latent representations to maximize the probability that code vectors are correctly matched in codebook. Finally,  the AFT module with dual-decoder architecture is introduced to flexibly supply information into the decoding process, which further improves the fidelity  of enhanced results while maintaining the perceptual quality. Extensive experiments demonstrate that our GLARE significantly outperforms the current SOTA methods on $5$ paired datasets and $4$ real-world datasets. The superior performance on low light object detection makes our GLARE an effective preprocessing tool in high-level vision tasks.


%
%
\bibliographystyle{splncs04}
\bibliography{main}

\clearpage

\title{Supplementary Material for GLARE: Low Light Image Enhancement via Generative Latent Feature based Codebook Retrieval} 

\titlerunning{GLARE: LLIE via Generative Latent Feature based Codebook Retrieval}

\author{Han Zhou\inst{1,*}\orcidlink{0000-0001-7650-0755} \and
Wei Dong\inst{1,*}\orcidlink{0000-0001-6109-5099} \and
Xiaohong Liu\inst{2,\dagger}\orcidlink{0000-0001-6377-4730} \and
Shuaicheng Liu\inst{3}\orcidlink{0000-0002-8815-5335} \and
Xiongkuo Min\inst{2}\orcidlink{0000-0001-5693-0416} \and
Guangtao Zhai\inst{2}\orcidlink{0000-0001-8165-9322} \and
Jun Chen\inst{1,\dagger}\orcidlink{0000-0002-8084-9332}}


\authorrunning{H.~Zhou et al.}


\institute{$^1$ McMaster University, $^2$ Shanghai Jiao Tong University, \\ $^3$ University of Electronic Science and Technology of China \\
\email{\{zhouh115, dongw22, chenjun\}@mcmaster.ca} \quad \email{liushuaicheng@uestc.edu.cn} \\
\email{\{xiaohongliu, minxiongkuo, zhaiguangtao\}@sjtu.edu.cn} \\ 
$^*$Equal Contribution \quad \quad $^{\dagger}$Corresponding Authors}

\maketitle

\ \\
\ \\
In this supplementary material, we provide detailed description for GLARE architecture and training objectives and discuss the potential limitation of GLARE.

\newcommand\appendixsectionformat{\titleformat{\section}{\normalfont\Large\bfseries}{\thesection}{1em}{}}
\renewcommand\thesection{\Alph{section}}

\section{Network Details of GLARE}
\label{sec:supp_glare_network}
\SubSection{Stage I: Normal-Light Codebook Learning } \label{sec:supp_codebook_dict}
\Paragraph{Architecture Detail} The NL encoder in VQGAN comprises 2 downsampling layers and the NL decoder has 2 upsampling operations. There are 2 residual blocks~\cite{residualblock} at each resolution level and 2 attention blocks at the resolution of $[\frac{W}{4}, \frac{H}{4}]$, where $W$ and $H$ represent the image width and height, respectively. The learnable codebook contains $1000$ discrete vector and the dimension for each vector is $3$.

\Paragraph{Training Objective} To establish a comprehensive codebook, the reconstruction loss $\mathcal{L}_{rec}$, the adversarial loss $\mathcal{L}_{adv}$, the codebook loss $\mathcal{L}_{code}$, and the perceptual loss $\mathcal{L}_{per}$~\cite{percep2018}  are introduced for training:
\vspace{-1mm}
\begin{equation}
\begin{split}
&\mathcal{L}_{rec} = \left \|  \mbf{I}_{nl} -  \mbf{I}^{rec}_{nl} \right \|_1, \\
&\mathcal{L}_{adv} =   -  \mathrm{log} \big(D\big( \mbf{I}^{rec}_{nl}\big)\big), \\
&\mathcal{L}_{code} = \left \| sg\small( \mbf{z}_{nl}\small) -\mbf{z}_{q} \right \|^2_2 + \beta\left \| sg\small( \mbf{z}_{q}\small) -\mbf{z}_{nl} \right \|^2_2, \\
&\mathcal{L}_{per} = \left \| \phi\big(\mbf{I}_{nl}\big) - \phi\big(\mbf{I}^{rec}_{nl}\big) \right \|^2_2,
\end{split}
\vspace{-3mm}
\label{eq_L_rec_adv_code}
\end{equation}
where $D$ is the discriminator~\cite{vq2017dis} and $\phi$ denotes the VGG19~\cite{vgg2015} feature extractor. Besides, $sg(\cdot)$ denotes the straight-through gradient estimator for facilitating the backpropagation via the non-differentiable quantization process, and $\beta$ is a pre-defined hyper-parameter. Moreover, we also leverage the multi-scale structure similarity loss $\mathcal{L}_{ssim}$~\cite{breakhaze2023han} and the latent semantic loss $\mathcal{L}_{sem}$~\cite{ridcp2023}, due to their outstanding performance in reconstructing image details. Therefore, the complete training loss utilized to learn the codebook is:
\vspace{-1mm}
\begin{equation}
\begin{split}
\mathcal{L}_{total} &= \mathcal{L}_{rec} + \lambda_{adv}\cdot \mathcal{L}_{adv} + \lambda_{code}\cdot \mathcal{L}_{code}\\
&+ \lambda_{per}\cdot \mathcal{L}_{per} + \lambda_{ssim}\cdot \mathcal{L}_{ssim} +\lambda_{sem}\cdot\mathcal{L}_{sem},
\end{split}
\vspace{-3mm}
\label{eq_supp_L_stage1}
\end{equation}
where $\lambda_{adv} = 0.0005$, $\lambda_{code} = 1$, $\lambda_{per} = 0.01$, $\lambda_{ssim}=0.2$, and $\lambda_{sem}= 0.1$ are coefficients for corresponding loss functions.

\SubSection{Stage II: Generative Latent Feature Learning} \label{sec:supp_normalization_flow}
\Paragraph{Architecture Details} Tab.~\ref{table_supp_architecture_stage2} outlines the architecture details of the conditional encoder $ E_{c}$ and I-LNF module $f_{\bm{\theta}}$. Basically, the conditional encoder contains a NL encoder and a simple convolution, and our I-LNF module consists of two flow layers, each sharing the same architecture as LLFlow~\cite{llflow2022}. Different from LLFlow, we remove all squeeze layers in our GLARE considering that our I-LNF operates at the feature level rather than image space.
\\
\Paragraph{Training Objective} We optimize the conditional encoder and I-LNF by minimizing the Negative Log-Likelihood (NLL) described in Eq. 3 of the manuscript. We provide more information about the training objective of stage II here. We first divide the invertible $f_{\bm{\theta}}$ into a sequence of $N$ invertible layers $\{f^{1}_{\bm{\theta}}, f^{2}_{\bm{\theta}}, ..., f^{N}_{\bm{\theta}}\}$. Then we utilize the following equation to demonstrate the feature flow for each layer:
\vspace{-2mm}
\begin{equation}
\mbf{h}^{i} = f^{i}_{\bm{\theta}}(\mbf{h}^{i-1}; \mbf{c}_{ll}), 
\vspace{-1mm}
\label{eq_supp_feat_flow}
\end{equation}
where $i\in [1, N]$, $\mbf{h}^{i-1}$ and $\mbf{h}^{i}$ represent the input and output of $ f^{i}_{\bm{\theta}}$ respectively. Specifically, $\mbf{h}^{0} = \mbf{z}_{nl}$ and $\mbf{h}^{N} = \mbf{v} $. Therefore, the detailed NLL loss can be expressed as:
\vspace{-1mm}
\begin{equation}
\begin{split}
\mathcal{L}
&(\bm\theta; \mbf{c}_{ll}, \mbf{z}_{nl} ) = -\mathrm{log} p_{\mbf{z}_{nl}|\mbf{c}_{ll} }(\mbf{z}_{nl}|\mbf{c}_{ll}, \bm\theta)\\
&= -\mathrm{log} p_{\mbf{v}}(\mbf{v}) - \sum_{i=1}^{N} \mathrm{log}|det \frac{\partial f^{i}_{\bm{\theta}}}{\partial \mbf{h}^{i-1}}(\mbf{h}^{i-1}; \mbf{c}_{ll})|.
\vspace{-2mm}
\end{split}
\label{eq_supp_nll_detailed}
\end{equation}

\renewcommand{\arraystretch}{1.2}

\begin{table}[t]
\centering
\captionsetup{font= small}
\caption{Architecture details of conditional encoder and I-LNF module. We design two flow layers and for each flow layer, we utilize the same architecture as LLFlow~\cite{llflow2022} and then remove all squeeze operations. With the condition of $\mbf{c}_{ll}$, the NL feature $\mbf{z}_{nl}$ is transformed to a latent feature $\mbf{v}$ with the mean of $\mbf{z}_{ll}$. [Key: ActNorm2d, InvertibleConv, and CondAffineCoupling: please refer to SRFlow~\cite{srflow2020} and the supplementary material of LLFlow~\cite{llflow2022} for a detailed description] }
    \begin{subtable}[t]{0.8\linewidth}
    \captionsetup{font= small}
    \caption{Architecture details of conditional encoder, cond\_conv and feat\_conv are simple convolutions to generate $\mbf{c}_{ll}$ and $\mbf{z}_{ll}$, respectively.}
    \centering
    \resizebox{\linewidth}{!}{
    \begin{tabular}{cc|c|c}
\hline
\multicolumn{2}{c|}{\textbf{Layers}}&\textbf{Configurations}&\textbf{Output Size} \\ 
\hline
\multicolumn{2}{c|}{Input}& Image Tensor & $3 \times W \times H$\\
\hline
\multirow{2}{*}{\begin{tabular}{c}
    Conditional\\
    Encoder
\end{tabular}}&\multicolumn{1}{|c|}{NL Encoder} &\multicolumn{1}{c|}{$c=128, k=3, s=1$}&\multicolumn{1}{c}{$3 \times \frac{W}{4} \times \frac{H}{4}$} \\ \cline{2-4}
&\multicolumn{1}{|c|}{cond\_conv} &\multicolumn{1}{c|}{$c=64,k=3, s=1$}&\multicolumn{1}{c}{$\mbf{c}_{ll}: 64 \times \frac{W}{4} \times \frac{H}{4}$} \\
\hline
\multicolumn{2}{c|}{feat\_conv} &\multicolumn{1}{c|}{$c=3,k=3, s=1$}&\multicolumn{1}{c}{$\mbf{z}_{ll}: 3 \times \frac{W}{4} \times \frac{H}{4}$}\\
\hline
\end{tabular}
    }    \label{table_supp_architecture_stage2_conditional_encoder}
    \end{subtable}    
    \begin{subtable}[t]{0.8\linewidth}
    \vspace{5mm}
    \centering
    \captionsetup{font= small}
    \caption{Architecture details of I-LNF.}
    \resizebox{0.9\linewidth}{!}{
    \begin{tabular}{c|c|c}
\hline
\multicolumn{1}{c|}{\textbf{Layers}}& \textbf{Configurations}&\textbf{Output Size} \\ 
\hline
\multirow{3}{*}{Input}& \multirow{3}{*}{Three latent tensors} & $ \mbf{z}_{nl}: 3 \times \frac{W}{4} \times \frac{H}{4}$ \\ \cline{3-3}
&& $ \mbf{c}_{ll}: 64 \times \frac{W}{4} \times \frac{H}{4}$\\ \cline{3-3}
&& $ \mbf{z}_{ll}: 3 \times \frac{W}{4} \times \frac{H}{4}$\\ \cline{3-3}
\hline
\renewcommand{\arraystretch}{1}
\multirow{3}{*}{FlowLayer1} &\multirow{3}{*}{\begin{tabular}{c}

            ActNorm2d \\
            InvertibleConv\\
            CondAffineCoupling$\times 12$
        \end{tabular}
} &\multirow{3}{*}{ $3\times \frac{W}{4} \times \frac{H}{4}$} \\
&&\\
&&\\
\hline
\multirow{3}{*}{FlowLayer2} &\multirow{3}{*}{\begin{tabular}{c}

            ActNorm2d \\
            InvertibleConv\\
            CondAffineCoupling$\times 12$
        \end{tabular}
} &\multirow{3}{*}{$\mbf{v}: 3 \times \frac{W}{4} \times \frac{H}{4}$} \\
&&\\
&&\\
\hline
\end{tabular}
    }
    \label{table_supp_architecture_stage2_I-LNF}
    \end{subtable}
\label{table_supp_architecture_stage2}    
\end{table}

\renewcommand{\arraystretch}{1}

\SubSection{Stage III: Adaptive Feature Transformation} \label{sec:supp_adptive_fusion}
\Paragraph{Training Objective} To train the AFT module proposed in stage III, we adopt multiple loss functions for optimization: $\mathcal{L}_{1}$ loss, multi-scale structure similarity loss $\mathcal{L}_{ssim}$~\cite{breakhaze2023han}, and perceptual loss $\mathcal{L}_{per}$~\cite{percep2018}. Therefore, the total loss function of Stage III can be formulated as:
\vspace{-1mm}
\begin{equation}
\mathcal{L}_{total} = \mathcal{L}_{1} + \lambda_{ssim}\cdot \mathcal{L}_{ssim} + \lambda_{per}\cdot \mathcal{L}_{per},
\vspace{-3mm}
\label{eq_supp_L_stage3}
\end{equation}
where $\lambda_{ssim}=0.2$ and $\lambda_{per}=0.01$.

\section{Limitation}
\label{sec:supp_limitation}
While extensive experiments demonstrate that our GLARE significantly outperforms the current SOTA methods in  LLIE and is verified to be an effective pre-processing method in low-light object detection task, our GLARE still presents several avenues for further exploration and refinement. While the VQGAN in our GLARE provides a highly important NL codebook, it comes at the cost of the efficiency of GLARE. Therefore, the efficiency of our GLARE requires further improvement for practical applications. Besides, only one generative method (normalizing flow) is studied in our work. To better demonstrate the generalizable capability of our GLARE, the task of employing other generative methods in our GLARE, such as diffusion model~\cite{diffusion}, requires further exploration.

\end{document}